\documentclass[runningheads]{llncs}

\usepackage[T1]{fontenc}
\usepackage{graphicx}
\usepackage{amsmath}
\usepackage{amssymb}
\usepackage{booktabs}
\usepackage{xcolor}
\usepackage[hidelinks]{hyperref}
\usepackage{xurl}
\graphicspath{{figures/}}

\begin{document}

\title{Behavioral Cloning Outperforms Entropy-Regularized RL: Critic-Driven Failure of Actor-Critic Methods on Adaptive Tumor Treatment}
\titlerunning{BC Outperforms Entropy-Regularized RL on Adaptive Tumor Treatment}

\author{Aleksandar Dimitrov\inst{1} \and Giacomo Spigler\inst{1}}
\authorrunning{A. Dimitrov and G. Spigler}
\institute{Department of Cognitive Science \& Artificial Intelligence,\\
           Tilburg University, Tilburg, The Netherlands\\
           \email{aleksandar.dimitrov10.2004@gmail.com, g.spigler@tilburguniversity.edu}}

\maketitle

\begin{abstract}

Adaptive dosing requires policies that reduce tumor burden without excessive toxicity. Learned dosing policies are typically judged against historical or heuristic comparators, which cannot show whether a policy has found the best behavior available. We instead study a three-population tumor-control ODE in which optimal-control analysis fixes the \emph{form} of a good schedule ---bang-bang dosing punctuated by a singular arc--- and construct a numerical controller of that form as a proxy for near-optimal behavior. Judged against this reference under a sustained-cure criterion ---200 consecutive days below 5\% carrying capacity--- Soft Actor-Critic (SAC) trained from scratch never reaches cure. Behavioral cloning (BC) of the reference reproduces it (100\% sustained cure, 30/30 seeds), but SAC fine-tuning of the cloned policy destroys it across five entropy coefficients, and TD3 and BC-regularized SAC fail identically; the pattern persists under multiplicative pharmacokinetic action noise. Along curative trajectories the post-collapse critic ranks the collapsed-policy action above the reference action in 96\% of states, concentrated in the maintenance phase, and the policy settles into a non-curative adaptive-therapy equilibrium. The reference is what makes this legible: against a heuristic comparator the fine-tuned policy would read as a competent controller rather than a failure.

\keywords{Reinforcement learning \and Behavioral cloning \and Actor-critic methods \and Soft Actor-Critic \and Optimal control \and Adaptive therapy \and Tumor control}

\end{abstract}

\section{Introduction}\label{sec:intro}

Treating cancer with cytotoxic chemotherapy can be highly effective, but often involves solving a complex optimization problem to determine the best dosing schedule. The standard maximum tolerated dose (MTD) protocol administers the highest dose a patient can tolerate, maximizing the immediate killing of tumor cell. Yet, MTD predictably triggers \emph{competitive release}: eliminating the drug-sensitive cells removes the ecological competition that had held a smaller population of drug-resistant cells in check, letting the resistant subpopulation expand and ultimately limiting durable response~\cite{gatenby2009adaptive}. \emph{Adaptive therapy}~\cite{zhang2022evolution} offers an alternative paradigm: rather than maximizing cell kill, it modulates the dose to preserve a residual population of sensitive cells whose competition suppresses the resistant cells, trading immediate tumor reduction for long-term control.

Deep reinforcement learning (RL) has prompted interest in autonomous adaptive-dosing controllers. RL has been applied to treatment scheduling in HIV, sepsis, and chemotherapy \cite{ernst2006clinical,komorowski2018artificial,yauney2018reinforcement}. In these settings, learned policies are typically evaluated against historical data or heuristic comparators rather than a reference controller derived from optimal-control analysis. Such evaluations can demonstrate relative improvement, but cannot establish whether a learned policy recovers a near-optimal schedule.

Here, we investigate a tumor-control problem in which optimal-control analysis indicates the \emph{form} of a good dosing schedule~\cite{pontryagin1962mathematical}: a \emph{bang-bang} schedule (alternating between zero and maximum dose) punctuated by a \emph{singular arc} (a precise intermediate maintenance dose). We construct a numerical controller of this form as a proxy for near-optimal behavior, and ask whether a Soft Actor-Critic agent can match it and, if not, why it fails.



We find that standalone SAC fails to discover curative trajectories despite extensive reward shaping. \emph{Behavioral cloning} (BC)---supervised imitation of an expert policy---from a Pontryagin-derived teacher achieves a 100\% durable cure rate, but subsequent SAC fine-tuning uniformly destabilizes it across all tested entropy coefficients. TD3~\cite{fujimoto2018td3} and DAPG-style BC-regularized SAC~\cite{rajeswaran2018learning} (which explicitly penalizes departure from the cloned policy) exhibit the identical collapse, isolating critic-driven gradient errors as an intervention-resistant failure mechanism. Probing the collapsed critic directly, we find it values the non-curative maintenance action above the curative action in 96\% of states along the curative trajectory, converting the critic-driven account from an inference into a direct measurement. The destabilized policy independently rediscovers the adaptive-therapy paradigm of Gatenby et al.~\cite{gatenby2009adaptive}, settling into a chronic-management equilibrium that never transitions to the cure basin.

Our contributions are threefold: (1)~a \emph{sustained-cure metric} that exposes specification gaming in standard tumor-RL cure criteria; (2)~an empirical characterization of when behavioral cloning suffices versus when actor-critic fine-tuning is appropriate for medical optimal control; and (3)~a mechanistic identification of critic-driven destabilization as the failure mode of modern actor-critic methods on problems with a known optimal-control structure, demonstrated directly via a critic Q-value analysis. All code is released Open Source at \url{https://github.com/AleksandarDimitrov10/entropy-regularized-rl-tumor-control}.


\section{Related Work}\label{sec:related}

\paragraph{Optimal control and adaptive therapy.} Mathematical models of tumor growth and treatment response have a long history in computational oncology~\cite{hahnfeldt1999tumor}. Adaptive therapy exploits competition between drug-sensitive and resistant populations to suppress resistance rather than continuously maximize cell kill~\cite{gatenby2009adaptive,strobl2022spatial}; it has also been evaluated in a small prostate-cancer study~\cite{zhang2022evolution}. Optimal-control analysis, particularly Pontryagin's Minimum Principle~\cite{pontryagin1962mathematical}, can yield bang-bang controls and, in some models, singular arcs~\cite{schattler2015optimal}. We use this control structure to construct the three-phase curative reference policy for our experiments.

\paragraph{RL for treatment scheduling.} RL has been used for treatment scheduling in HIV, sepsis, and chemotherapy~\cite{ernst2006clinical,komorowski2018artificial,yauney2018reinforcement}. These studies commonly evaluate learned policies against historical, clinical, or heuristic comparators rather than a reference controller derived from optimal-control analysis. Such comparisons establish relative performance, but cannot test whether a learned policy recovers a near-optimal schedule of known structure --- the question considered here.

\paragraph{Actor-critic fine-tuning.} Soft Actor-Critic (SAC) is an entropy-regularized off-policy actor-critic method~\cite{haarnoja2018soft}; maximum-entropy RL also has a formal robustness interpretation for certain disturbances to dynamics and rewards~\cite{eysenbach2021maxent}. TD3 removes the entropy term while retaining twin critics~\cite{fujimoto2018td3}, whereas DAPG-style methods use demonstrations to regularize policy learning~\cite{rajeswaran2018learning}. Because deep-RL results can be seed-sensitive~\cite{henderson2018deep}, we evaluate multiple seeds and algorithmic variants. Recent offline-to-online methods such as RLPD and Cal-QL~\cite{ball2023efficient,nakamoto2023calql} are relevant future baselines, but are outside the scope of this study.

\section{Methods}\label{sec:methods}

\subsection{Environment and sustained-cure metric}

We develop \textsc{TumorEnvV2}, a three-population ordinary differential equation (ODE) model tracking sensitive cells $S$, resistant cells $R$, and accumulated drug toxicity $T$ under Lotka-Volterra competition with selective drug action:
\begin{align}
\dot{S} &= r_S\, S\Big(1 - \tfrac{S + \alpha_{RS} R}{K}\Big) - d_S\, S - \gamma_S\, u\, S, \\
\dot{R} &= r_R\, R\Big(1 - \tfrac{R + \alpha_{SR} S}{K}\Big) - d_R\, R + \mu_{\mathrm{base}}\, S, \\
\dot{T} &= \alpha_{\mathrm{tox}}\, u - \beta_{\mathrm{tox}}\, T,
\end{align}
where $u \in [0,1]$ is the prescribed dose and $K$ is the carrying capacity. The drug is selective ($\gamma_R = 0$), modeling acquired resistance, and a fraction $\mu_{\mathrm{base}}$ of sensitive cells mutates to resistant cells per step. Total burden is $V = S + R$, with cure threshold $V/K < 0.05$. We set $\alpha_{\mathrm{tox}} = 0.08$ (from a base value of $0.10$) to open a non-degenerate cure/toxicity frontier; MTD remains 100\% toxic across 30 seeds under this change. The agent observes the six-dimensional vector
\begin{equation*}
\big(\log_{10}(S{+}1),\ \log_{10}(R{+}1),\ \log_{10}(V{+}1),\ T/T_{\max},\ t/t_{\max},\ c/c_{\mathrm{sustain}}\big),
\end{equation*}
where $c$ counts consecutive days below the cure threshold. Logarithmic scaling is used because cell counts span five orders of magnitude.

\paragraph{Sustained-cure metric.} The standard criterion (5 consecutive days below $5\%\,K$) can be satisfied by a transient crash of the sensitive population. We therefore require \emph{200 consecutive days} below $5\%\,K$ with resistant cells a minority of residual tumor ($R < 0.5\,(S{+}R)$), corresponding to roughly 6.5 months of stable control and matching adaptive-therapy trial follow-up windows~\cite{zhang2022evolution}.

\subsection{Pontryagin-derived teacher and behavioral cloning}

\paragraph{Teacher.} We construct a three-phase teacher (\textsc{SmartProxy\,v3}) approximating the optimal-control structure: \emph{(i)}~wait at $u = 0$ until $R < R_{\mathrm{gate}} = 10^5$, letting competition and metabolic cost deplete the resistant population; \emph{(ii)}~crash at $u = 1$, dosing only while predicted next-step toxicity remains below $T_{\max}$; and \emph{(iii)}~hold the singular-arc maintenance dose $u = 0.25$, at which $S$ equilibrates near $4$--$5\%\,K$ with $T_{\mathrm{eq}} = 1.33 < T_{\max} = 1.5$. A regrowth guard reverts to $u = 1$ near the threshold. Both $u = 0.25$ and $R_{\mathrm{gate}} = 10^5$ were fixed by coarse validation sweeps. We emphasize that this controller is not a certified optimum: its structure follows the form that optimal-control analysis yields for models of this class~\cite{schattler2015optimal}, but its constants were tuned rather than derived. In a setting this simplified such a numerical approximation can be close to optimal, and we use it as a \emph{proxy} for the optimal policy --- a reference for the best behavior we can currently identify, rather than a proven upper bound.

\paragraph{Behavioral cloning.} We collect 500 teacher demonstration episodes ($\approx$158k transitions, all curative) and train an MLP actor ($[256,256]$, identical in architecture to the SAC actor) for 30k gradient steps to minimize $\mathbb{E}_{(s,a^*)\sim\mathcal{D}}\lVert \pi_\theta(s) - a^*\rVert_2^2$, evaluating on 20 held-out seeds. Distilling the teacher into the SAC actor's architecture allows us to attribute changes under fine-tuning to policy weights rather than to a difference in representation.

\subsection{SAC fine-tuning and interventions}

\paragraph{Fine-tuning.} We fine-tune the BC actor with SAC~\cite{haarnoja2018soft} using a learning rate of $3\times10^{-5}$ (to preserve BC weights), batch size 512, replay buffer $2\times10^5$, $\gamma = 0.999$, $\tau = 0.005$, 5{,}000 learning-start steps, and $[256,256]$ networks. The reward combines cure-zone proximity and resistance-suppression bonuses with burden, drug, gated-toxicity, and dose-smoothness penalties, plus a terminal cure bonus ($+10^4$) and failure penalty ($-500$); terminal events dominate per-step reward magnitudes. Pilot runs with linearly rescaled weights reproduced the same collapse, suggesting that it depends on reward shape rather than overall scale.

\paragraph{Interventions.} To probe the source of the destabilization, we run four interventions: \emph{(a)}~an entropy-coefficient sweep $\alpha \in \{0,\,10^{-4},\,10^{-3},\,10^{-2},\,10^{-1}\}$ over three seeds; \emph{(b)}~fine-tuning with TD3~\cite{fujimoto2018td3}, a deterministic actor-critic baseline with twin critics and no entropy term; \emph{(c)}~DAPG-style~\cite{rajeswaran2018learning} BC regularization, adding $\beta\lVert\pi_\theta(s) - \pi_{\mathrm{BC}}(s)\rVert_2^2$ to the actor loss for $\beta \in \{0.1, 1.0, 10.0\}$ (with $\alpha = 0.001$); and \emph{(d)}~the entropy sweep in a stochastic environment with multiplicative pharmacokinetic action noise $u_{\mathrm{actual}} = u\cdot\mathcal{N}(1,\sigma^2)$, $\sigma = 0.20$, applied at every training and evaluation step.

\paragraph{SAC from scratch.} As a reference point without demonstrations, we also
train SAC from random initialization on the same environment and reward. To avoid
handicapping it we use the Stable-Baselines3 default learning rate of
$3\times10^{-4}$ rather than the $3\times10^{-5}$ used for fine-tuning (chosen there
to preserve BC weights), and a budget of 100k steps --- more than three times that of
the fine-tuning experiments --- over three seeds.

\paragraph{Robustness evaluations.} We stress-test the teacher under multiplicative PK noise (up to $\sigma = 1.5$) and randomized initial conditions ($T_0 \in [0, 0.4]$, $R_0 \in [2\times10^7, 10^8]$). We also test the teacher and BC-only policy under Gaussian observation noise added to the state components at $\sigma \in \{0.05, 0.10, 0.20, 0.30, 0.50\}$; each condition uses 20 episodes.

\subsection{Critic Q-value analysis}
To examine the critic-driven hypothesis, we take a collapsed BC+SAC checkpoint, independently verified to achieve no cures in 30 held-out evaluation episodes. Along the teacher's curative trajectory, we query its critic for the teacher action $a_{\mathrm{curative}}$ and the collapsed policy action $a_{\mathrm{collapsed}}$ at each visited state, taking the minimum of the twin-critic estimates. Stable-Baselines3~\cite{raffin2021stablebaselines} trains the critic on actions scaled to $[-1,1]$, rather than the environment range $[0,1]$; we apply the same scaling before each query. Without it, the critic is evaluated off-distribution. We report the fraction of states for which $Q(s, a_{\mathrm{collapsed}}) > Q(s, a_{\mathrm{curative}})$ and repeat the analysis over ten randomized initial conditions ($T_0 \in [0, 0.4]$, $R_0 \in [2\times10^7, 10^8]$). Cure rates throughout use Clopper--Pearson exact 95\% binomial confidence intervals, and pairwise method comparisons use Fisher's exact test.

\section{Results}\label{sec:results}

\subsection{Behavioral cloning cures; actor-critic fine-tuning collapses}

We evaluate six policies on the sustained-cure metric across 30 fresh seeds (Table~\ref{tab:baselines}). The naive baselines fail entirely: MTD fails toxically on every seed and periodic 5/5 dosing times out, both at 0\% cure. The Pontryagin-derived teacher (SmartProxy~v3) and its behavioral-cloned student, evaluated at fine-tuning onset, both reach 100\% durable cure. After 30k fine-tuning updates the same run cures in none of the 30 seeds, every episode ending in timeout. Note that the collapsed policy attains the \emph{lowest} final resistant population of any method while never curing, because it holds the sensitive population near $30\%\,K$ and total burden never crosses the threshold. SAC trained from scratch on the same environment and reward never reaches the cure basin either: across three seeds no evaluation at any point during training recorded a cure, and the final policies leave the sensitive population near $82\%\,K$, i.e.\ they barely treat at all.

\begin{table}[t]
\centering
\caption{Sustained-cure metric ($n = 30$ seeds; 95\% Clopper-Pearson exact binomial confidence intervals). BC is evaluated at fine-tuning onset, before any gradient updates; the BC$+$SAC row is the same run after 30k updates.}
\label{tab:baselines}
\begin{tabular}{lcccc}
\toprule
Policy & Cure rate & 95\% CI & Final $S$ & Final $R$ \\
\midrule
MTD ($u = 1$)                      & 0\%   & [0, 12]\%   & $1.91\times10^{7}$ & $2.07\times10^{7}$ \\
Periodic 5/5                       & 0\%   & [0, 12]\%   & $2.38\times10^{7}$ & $2.47\times10^{7}$ \\
SAC from scratch                   & 0\%   & [0, 12]\%   & $8.24\times10^{8}$ & $2.18\times10^{4}$ \\
Teacher (\textsc{SmartProxy\,v3})  & 100\% & [88, 100]\% & $2.81\times10^{7}$ & $4.28\times10^{6}$ \\
BC, at fine-tuning onset           & 100\% & [88, 100]\% & $3.07\times10^{7}$ & $4.23\times10^{6}$ \\
BC $+$ SAC, after 30k updates      & 0\%   & [0, 12]\%   & $2.92\times10^{8}$ & $2.76\times10^{5}$ \\
\bottomrule
\end{tabular}
\end{table}

\begin{figure}[t]
  \centering
  \includegraphics[width=0.9\textwidth]{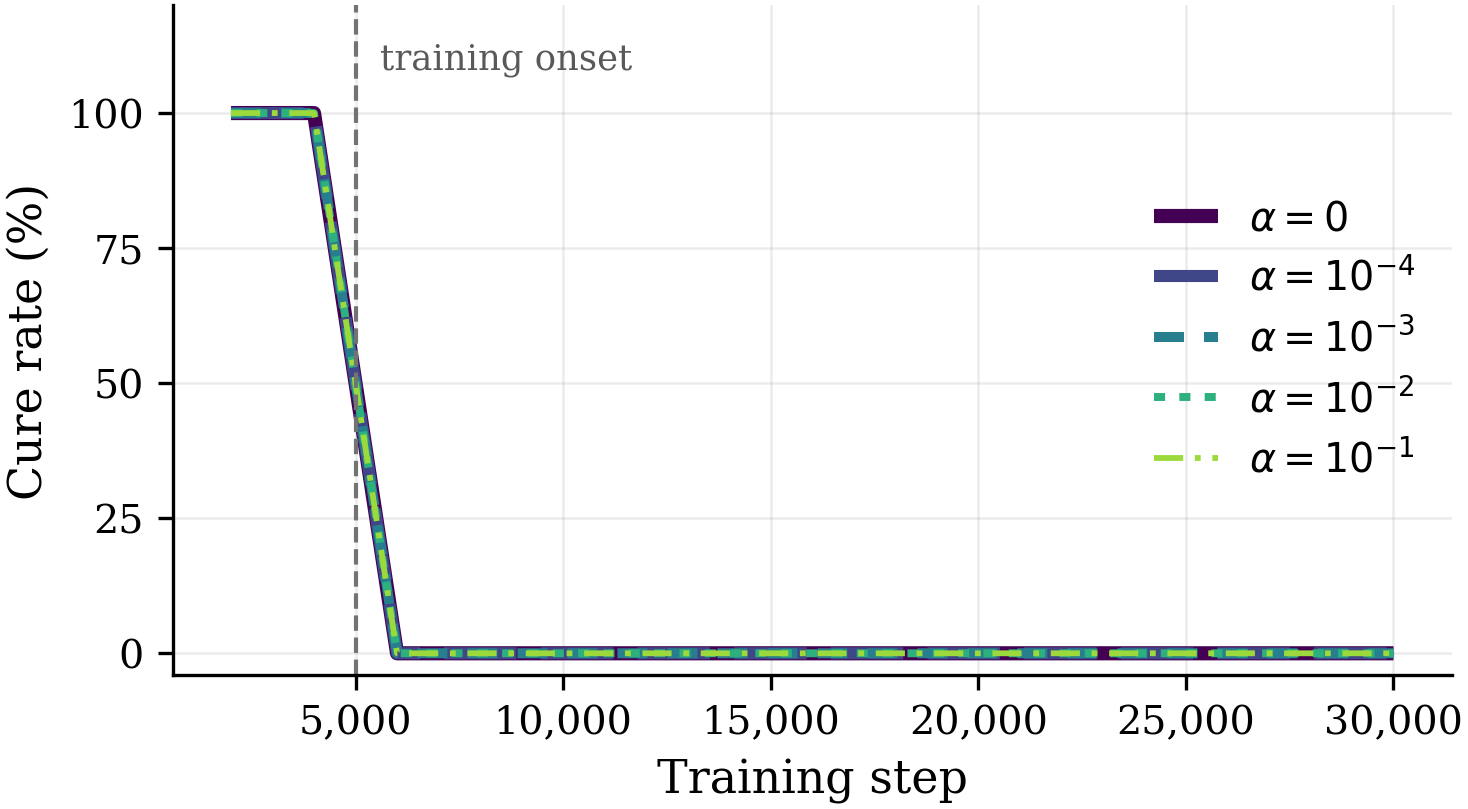}
  \caption{Entropy-coefficient sweep. All five $\alpha$ values show collapse from 100\% to 0\% cure within 1{,}000 gradient steps of training onset. The curves coincide exactly and are drawn with distinct dash patterns so that the overlap remains visible.}
  \label{fig:entropy}
\end{figure}

\paragraph{Varying entropy does not avert collapse.} Across 15 fine-tuning runs (five entropy coefficients $\times$ three seeds), every configuration exhibits the same cure dynamics: a 100\% peak at the first evaluation, followed by collapse to 0\% within 1{,}000 gradient steps of training onset (Fig.~\ref{fig:entropy}). Thus, in this experimental setup, changing $\alpha$ across four orders of magnitude does not prevent collapse.

\paragraph{Removing entropy and adding demonstration anchoring do not help.} TD3 ---a deterministic actor-critic method with twin critics and no entropy bonus--- also collapses (100\% $\to$ 0\% within 1{,}000 steps, 30 seeds). DAPG-style BC-regularized SAC, which adds the penalty $\beta\lVert\pi_\theta(s) - \pi_{\mathrm{BC}}(s)\rVert_2^2$ to the actor loss, likewise collapses for every $\beta \in \{0.1, 1.0, 10.0\}$. Even at $\beta = 10.0$, where the anchor term is comparable in magnitude to per-step rewards, the policy collapses within 1{,}000 updates. Together, these interventions show that collapse persists without entropy and despite explicit imitation regularization.

\paragraph{The collapse survives environment stochasticity.} Repeating the entropy sweep with multiplicative pharmacokinetic action noise ($\sigma = 0.20$) at every training and evaluation step does not rescue the policy. The noise-brittle BC initialization peaks at 40\% cure rather than 100\%, but still collapses to 0\% for all five $\alpha$ values and three seeds ($40 \to 0$, $n = 15$, zero variance). The destabilization therefore also occurs in the stochastic environment.

\subsection{Critic misvaluation along curative trajectories}\label{sec:criticqresults}

\begin{figure}[t]
  \centering
  \includegraphics[width=0.95\textwidth]{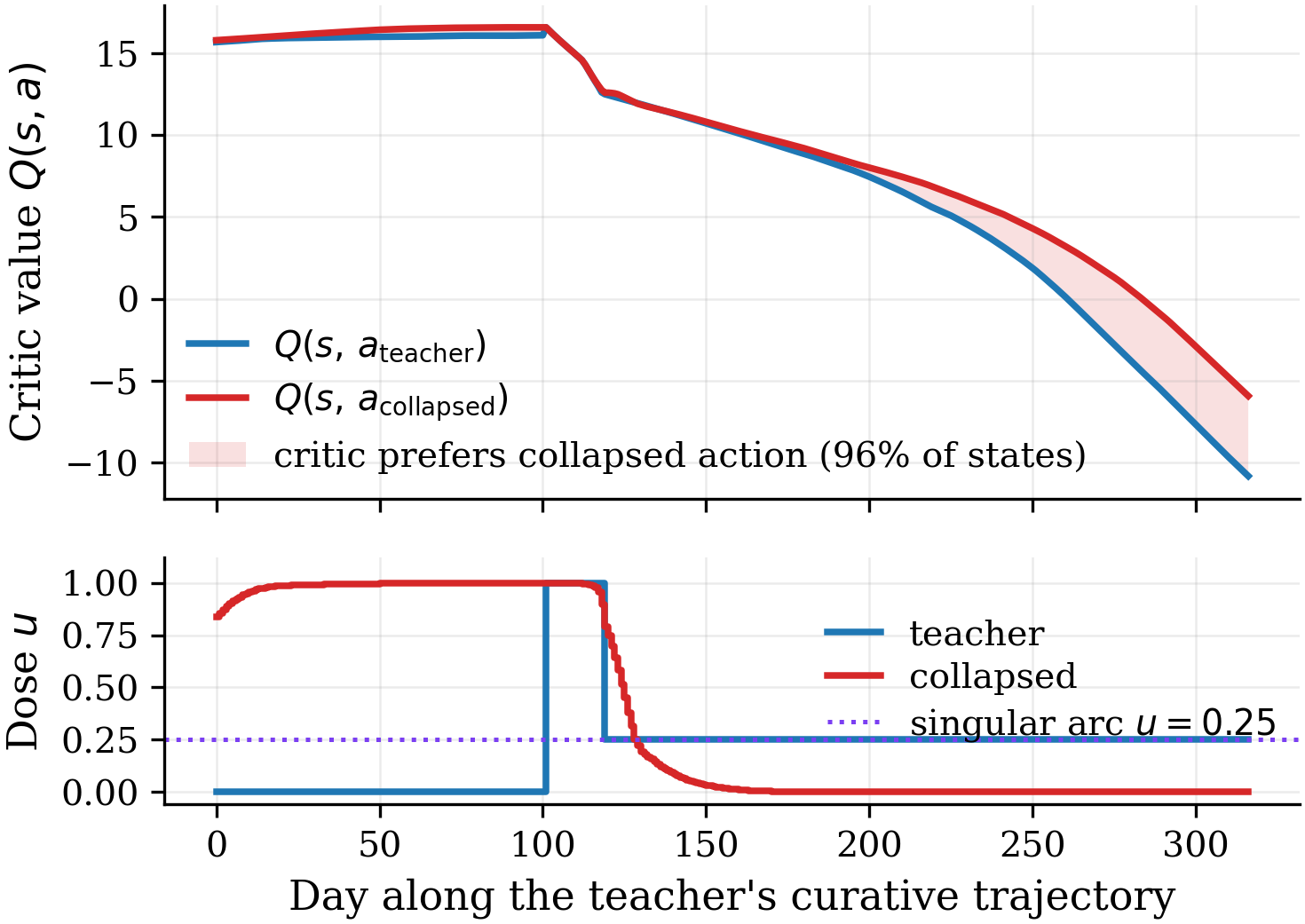}
  \caption{Along the teacher's curative trajectory, the collapsed critic assigns a higher estimated value to the collapsed-policy action than to the teacher action in 96\% of states (shaded), with the largest gap during the singular-arc maintenance phase.}
  \label{fig:criticq}
\end{figure}

If collapse involves critic misvaluation, the collapsed critic should assign a higher value to its own action than to the teacher action. Along the teacher's curative trajectory, we therefore query the collapsed critic for both actions, scaling them to $[-1,1]$ to match critic training. The critic assigns a higher estimated value to the collapsed-policy action in \textbf{96\% of states} (mean value gap $+1.26$, maximum $+4.89$; Fig.~\ref{fig:criticq}). Across ten randomized initial conditions, this holds in $94.8\% \pm 1.4\%$ of states (mean gap $+1.23 \pm 0.02$), while the teacher cures all ten. The gap is small in the drug-free waiting phase (mean $+0.35$) and larger in the singular-arc maintenance phase (mean $+1.83$), where the precise maintenance dose is required. These estimates provide direct evidence of critic mis-ranking along the curative trajectory and are consistent with critic-mediated policy drift.

\subsection{The failure is a maintenance trap, and BC does not inherit the teacher's robustness}

\begin{figure}[t]
  \centering
  \includegraphics[width=\textwidth]{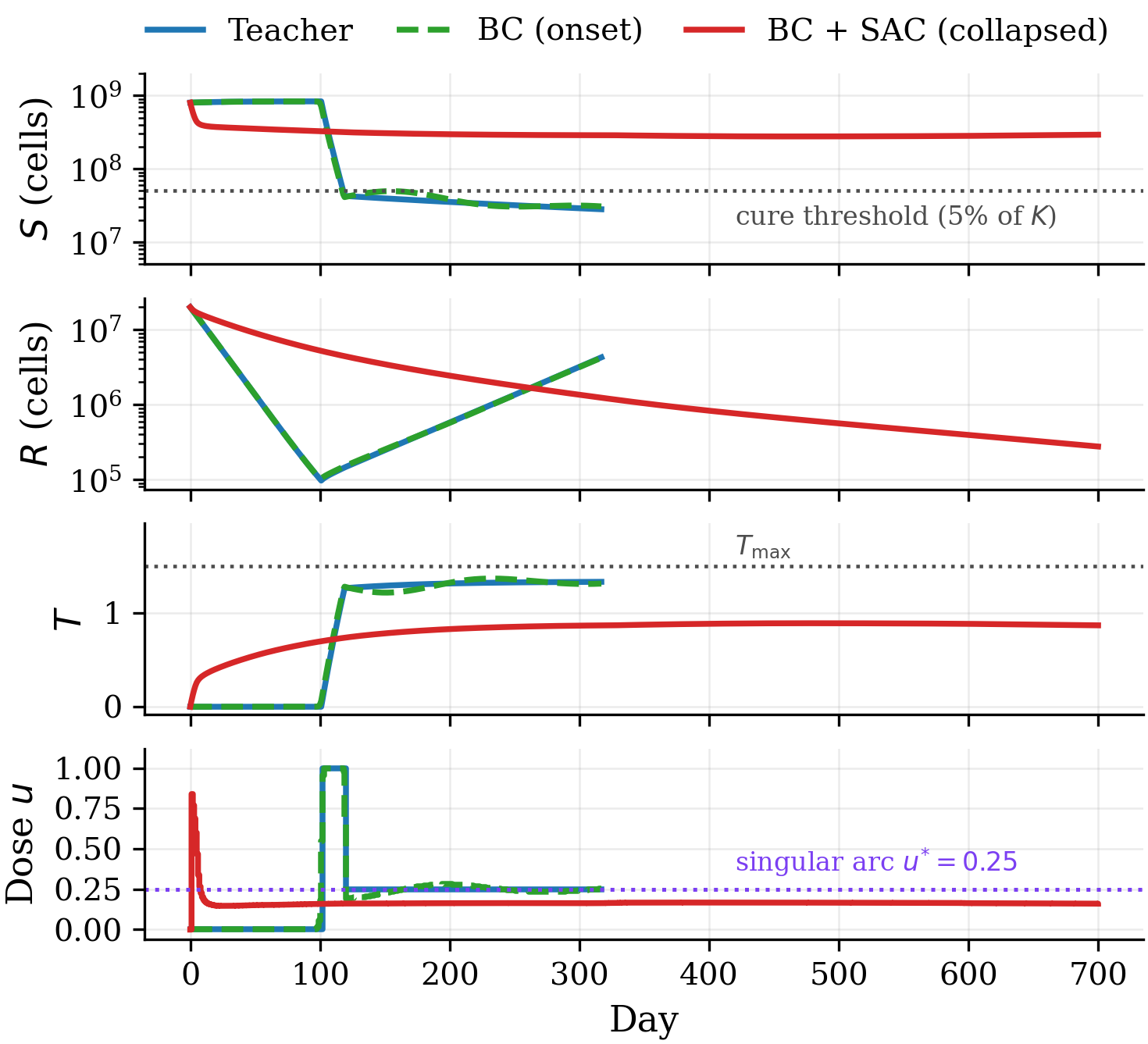}
  \caption{Failure mode of BC+SAC versus the teacher and BC-only policies from a shared initial condition. The collapsed policy holds a high-$S$/low-$R$ equilibrium but never enters the cure basin.}
  \label{fig:trace}
\end{figure}

Tracing the collapsed policy against the teacher and BC-only policy from a shared initial condition (Fig.~\ref{fig:trace}) illustrates the failure mode. The teacher and BC-only policies follow the wait-crash-maintain structure, whereas the collapsed BC+SAC policy issues a low non-zero dose from the first step and settles into a chronic-management equilibrium ($S \approx 30\%\,K$, $R \approx 3\times10^{5}$). It remains there for the full 700-day horizon without crossing the cure threshold, yielding an adaptive-therapy-like attractor~\cite{gatenby2009adaptive}.

Under Gaussian observation noise, the BC-only policy reaches 5\% cure at $\sigma = 0.05$ and 0\% beyond, whereas the rule-based teacher maintains 100\% cure up to $\sigma = 0.10$ and 90\% at $\sigma = 0.20$. Thus, despite equivalent clean-observation performance, the BC policy does not inherit the teacher's observation-noise robustness in this setting.

\subsection{Ablation overview}

Beyond the policies in Table~\ref{tab:baselines}, we also evaluated SAC with
additional reward shaping, an earlier teacher variant
(\textsc{SmartProxy\,v1}), BC$+$TD3, and BC$+$DAPG-SAC. Every actor-critic
configuration reached 0\% final sustained cure; only the teacher policies and BC
without fine-tuning cured.

\section{Discussion}\label{sec:discussion}

Our results differ from much of the RL-for-dosing literature, which commonly evaluates policies against historical or heuristic baselines~\cite{ernst2006clinical,komorowski2018artificial,yauney2018reinforcement}. Here, the curative reference policy exposes a more demanding failure mode: actor-critic fine-tuning converges to chronic tumor control rather than the curative trajectory. This distinction depends on having the reference. Against a historical or heuristic comparator the collapsed policy would not obviously read as a failure at all --- it suppresses resistance to the lowest level of any method we tested (Table~\ref{tab:baselines}) and holds the tumor stable for the full horizon. It is identifiable as a failure only because a near-optimal controller exists for this model, and because the sustained-cure criterion asks for durable control rather than a threshold crossing. The critic analysis associates the collapse with mis-ranking of the teacher and collapsed-policy actions, particularly near the singular arc. This evidence is consistent with, but does not by itself prove, critic-mediated policy drift.

BC is useful here as a diagnostic initialization rather than a deployment solution. It provides the SAC actor's architecture while preserving the teacher's behavior under clean observations, so subsequent changes can be attributed to fine-tuning rather than representation. However, it neither benefits from the tested fine-tuning procedures nor inherits the teacher's observation-noise robustness. The latter concerns observation perturbations, distinct from the dynamics and reward disturbances considered in maximum-entropy RL robustness results~\cite{eysenbach2021maxent}.

\paragraph{Adaptive-therapy interpretation.} The adaptive-therapy-like equilibrium provides an interpretable alternative to cure, but its clinical value depends on the treatment objective and model assumptions.

\paragraph{Practical implications.} Where a theory-derived controller is available and its assumptions are appropriate, it should be the reference against which learned policies are evaluated. Calibration-based offline-to-online methods, including RLPD and Cal-QL~\cite{ball2023efficient,nakamoto2023calql}, are relevant future baselines.

\paragraph{Limitations.} These results are from one ODE model and parameterization and should be read as evidence about this controlled setting, rather than as a general failure of actor--critic methods. We test SAC, TD3, and DAPG, with three training seeds per sweep condition; observed zero variance does not replace broader replication. The reference controller is a tuned approximation rather than a certified optimum, which bounds what we can claim about optimality. Testing further algorithms and model parameterizations, and eventually clinically informed settings, remains necessary.



\section{Conclusion}\label{sec:conclusion}

What makes this model system informative is that a near-optimal controller can be constructed for it, which turns ``did the agent do well?'' into a measurable question. Against that reference: SAC trained from scratch did not discover a curative policy; BC of the Pontryagin-derived teacher reproduced sustained cure under clean observations; and SAC, TD3, and BC-regularized SAC fine-tuning all collapsed to a non-curative maintenance regime. The collapsed critic ranked its own action above the teacher action along curative trajectories, especially near the maintenance phase, consistent with critic misvaluation. Because that collapsed policy is stable, low-resistance, and superficially plausible, it would plausibly pass an evaluation based on heuristic comparators alone. These results therefore motivate evaluating learned dosing policies against durable outcome criteria and, where available, theory-derived reference controllers rather than only historical or heuristic baselines.

\subsubsection*{Disclosure of Interests.} The authors have no competing interests to declare that are relevant to the content of this article.

\bibliographystyle{splncs04}
\bibliography{references}

\end{document}